\documentclass[10pt,twocolumn,letterpaper]{article}

\usepackage{iccv}              

\usepackage{amsmath}
\usepackage{amssymb}
\usepackage{booktabs}
\usepackage{algorithm}
\usepackage{algorithmic}
\usepackage{colortbl}
\usepackage{xcolor}
\usepackage{soul}
\definecolor{iccvblue}{rgb}{0.21,0.49,0.74}
\usepackage[pagebackref,breaklinks,colorlinks,allcolors=iccvblue]{hyperref}

\def\paperID{3} 
\def\confName{ICCV}
\def\confYear{2025}

\title{Automated Feature Tracking for Real-Time Kinematic Analysis and \\
Shape Estimation of Carbon Nanotube Growth}

\author{
Kaveh Safavigerdini$^{1}$ 
\and 
Ramakrishna Surya$^{1}$ 
\and 
Jaired Collins$^{1}$ 
\and
Prasad Calyam$^{1}$ 
\and
Filiz Bunyak$^{1}$ 
\and
Matthew R. Maschmann$^{1}$
\and
Kannappan Palaniappan$^{1}$
\and
$^{1}$University of Missouri-Columbia, USA\\
\tt\small \{ksgh2, rst7b, jrcqw5, calyamp, bunyak, maschmannm, pal\}@missouri.edu \\
}

\begin{document}
\maketitle
\begin{abstract}
Carbon nanotubes (CNTs) are critical building blocks in nanotechnology, yet the characterization of their dynamic growth is limited by the experimental challenges in nanoscale motion measurement using scanning electron microscopy (SEM) imaging. Existing ex situ methods offer only static analysis, while in situ techniques often require manual initialization and lack continuous per-particle trajectory decomposition. We present Visual Feature Tracking (VFTrack) an in-situ real-time particle tracking framework that automatically detects and tracks individual CNT particles in SEM image sequences. VFTrack integrates handcrafted or deep feature detectors and matchers within a particle tracking framework to enable kinematic analysis of CNT micropillar growth. A systematic using 13,540 manually annotated trajectories identifies the ALIKED detector with LightGlue matcher as an optimal combination (F1-score of 0.78, $\alpha$-score of 0.89). VFTrack motion vectors decomposed into axial growth, lateral drift, and oscillations, facilitate the calculation of heterogeneous regional growth rates and the reconstruction of evolving CNT pillar morphologies. This work enables advancement in automated nano-material characterization, bridging the gap between physics-based models and experimental observation to enable real-time optimization of CNT synthesis. Our source code is available at \url{https://github.com/kavehsfv/VFTrack}.
\end{abstract}    
\section{Introduction}
\label{sec:intro_n}
\textbf{Carbon Nanotubes (CNTs)} are pivotal in advancing nanotechnology across electronics, materials science \cite{moghanlou2025accelerated}, optics \cite{najmi2024thermal}, and polymer composites \cite{najmi2025effects, azizian2025optimization} due to their superior electrical conductivity~\cite{zhu2021embedding,safavigerdini2023predicting, hajilounezhad2021predicting}, mechanical robustness, and unique chemical properties~\cite{anzar2020carbon,choudhary2022contemporary,carter2017high}. These graphene-based tubular structures are synthesized as dense \textbf{CNT forest micropillars}~\cite{rasulev2020ecotoxicological, nguyen2023cnt}—vertically aligned microarchitectures exhibiting exceptional thermal/electrical conductivity, durability, and near-perfect optical absorption~\cite{surya2023using,safavigerdini2023creating, nguyen2022self}. Understanding CNT micropillar growth dynamics requires advanced monitoring techniques~\cite{hajilounezhad2019exploration, kotha2023deep}, categorized as \textbf{ex situ} methods analyzing static post-synthesis samples and \textbf{in situ} techniques monitoring real-time growth.
 
\begin{figure}[!t]
    \centering 
    \includegraphics[width=0.475\textwidth]{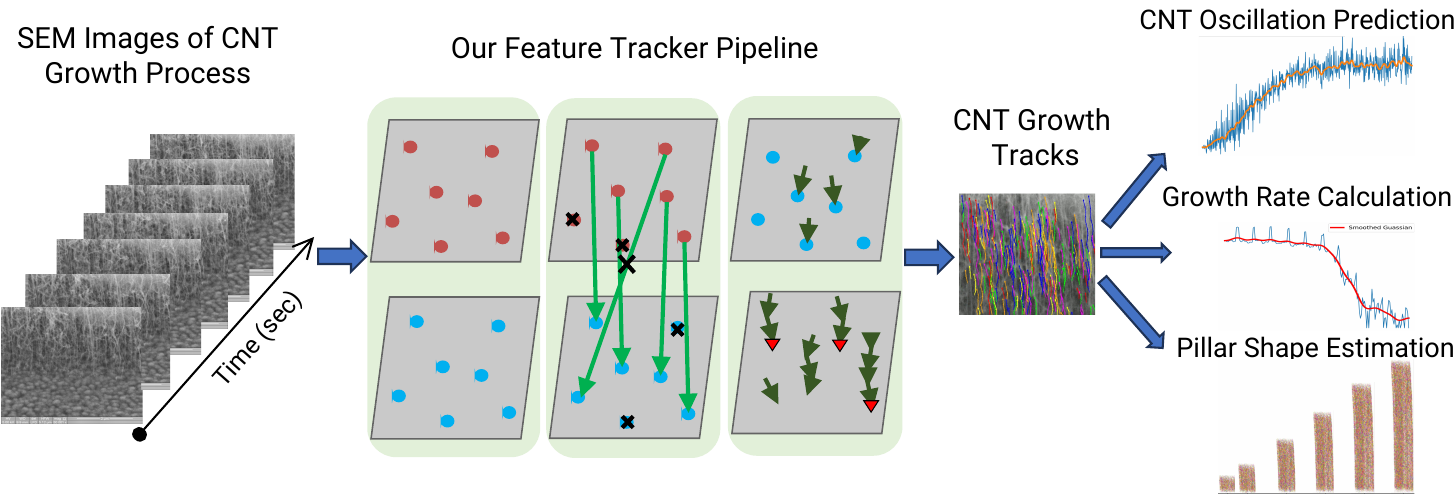}\\
    \caption{
    Overview of our VFTrack (particle) feature tracking pipeline that processes a sequence of SEM-based CNT growth images. Nanoscale CNT particles are automatically detected and tracked to quantify their oscillations and growth rates in real-time while reconstructing the evolving shape of the CNT micropillar.}\label{fig:method_input_output}
\end{figure}

\textit{Ex situ} mechanical characterization and static growth control techniques \textbf{fail to capture real-time CNT growth dynamics}. Shen et al. \cite{shen2020effect} employed nanoindentation and micropillar compression to link vertically aligned CNT orientation with anisotropic Young’s modulus, hardness, and fracture modes, but as an ex-situ technique, it cannot capture individual growth kinetics. Likewise, Zhou et al. \cite{zhou2010orthogonal} demonstrated static 3D architectures via orthogonal CVD growth, yet this approach offers no real-time view of morphological evolution or nanoscale fluctuations during synthesis. 

\textit{In situ} imaging approaches provide dynamic information but lack \textbf{continuous per-particle trajectory decomposition and oscillation analysis}. Adachi et al. \cite{adachi2024kinematics} used in situ TEM to observe electromigration-driven sliding of Co nanorods inside MWCNTs and extract force parameters but did not extend to continuous tracking of growing CNT ensembles. Surya et al. \cite{surya2024cnt, surya2024measurements} applied ESEM with digital image correlation and the Meta CoTracker algorithm to measure average forest growth rates under a rigid-body assumption, lacking per-particle trajectory decomposition or oscillation metrics. Pimonov et al. \cite{pimonov2021dynamic} demonstrated homodyne polarization microscopy for subsecond imaging of individual CNTs, revealing stochastic growth–pause–shrinkage switches but without mapping spatial drift or oscillation amplitudes in SEM frames. On the modeling side, Eltaher et al. \cite{eltaher2021dynamic} developed a doublet-based Timoshenko beam theory to predict CNT dynamic response under moving loads, yet this framework remains purely theoretical and disconnected from experimental image sequences.

Existing approaches to analyzing CNT kinematic properties are largely \textbf{manual and offline}, lacking automation and real-time responsiveness. Most current methods require manual seed selection and post-processing, preventing timely adjustments. In contrast, \textit{real-time CNT investigation} enables live monitoring and adaptive control of experimental conditions such as temperature and pressure, facilitating the attainment of desired mechanical properties during growth.

Conventional computer vision solutions such as \textbf{particle tracking} \cite{newby2018convolutional, shen2017single, liu2022situ}, \textbf{cell tracking} \cite{10159213, Bao_2021_ICCV, 9626982} and \textbf{point tracking}~\cite{karaev2024cotracker, doersch2023tapir, harley2022particle} face inherent limitations in CNT video analysis. Particle tracking algorithms typically need manual initialization of particle locations and exhibit performance degradation over long image sequences, while point tracking methods require initial point selections and recurrent reinitialization for new features across frames, hindering scalability and robustness.

To address these gaps, we propose a robust, real-time image-based tracking framework. Our approach is built on a systematic evaluation of state-of-the-art feature detector~\cite{barroso2022key, aktar2018performance} (\textit{SIFT~\cite{lowe2004distinctive}, ALIKE~\cite{zhao2023aliked}, DISK~\cite{tyszkiewicz2020disk}, SuperPoint~\cite{detone2018superpoint}}) and feature matcher \cite{ma2021image, schonberger2017comparative} (\textit{SuperGlue~\cite{sarlin2020superglue}, LightGlue~\cite{lindenberger2023lightglue}}). The framework is designed to automatically detect and match rich CNT particles across successive SEM frames, decompose their motions into growth, drift, and oscillation components, and reconstruct two-dimensional pillar morphologies. 

\textbf{The innovation of this study} lies in three key areas that overcome the aforementioned limitations. First, our framework provides \textbf{fully automated tracking without manual initialization}, unlike conventional point-tracking methods that require seed selection. Second, it enables \textbf{continuous per-particle trajectory decomposition}, moving beyond simple growth rate averages to simultaneously quantify axial growth, lateral drift, and orientation oscillations. Third, its computational efficiency facilitates \textbf{real-time analysis and control}, creating the potential for adaptive optimization of synthesis conditions. Using this pipeline, researchers can implement closed-loop control systems and apply different adaptive control theory models \cite{yaghooti2020adaptive, yaghooti2024stabilizing} directly to their SEM machines for adaptive synthesis optimization. By integrating these capabilities, our methodology bridges the critical gap between theoretical dynamic models and experimental observation, as illustrated in \cref{fig:method_input_output}, representing a significant advancement in automated nanomaterial characterization.
\section{Methodology}
\label{sec:methodology}
This section is organized into four primary parts. Initially, we describe SEM video acquisition of CNT growth under varied conditions; second, we introduce a novel feature-tracking algorithm for extracting CNT keypoint trajectories; third, we present a vector composition-based kinematic analysis to compute CNT displacements, orientations, growth rates, and oscillations; and fourth, we outline a pillar-shape estimation method that reconstructs CNT micropillar morphology by aggregating tracked movements over time.

\subsection{Preparing SEM Videos of CNT Images}
\label{subsec:preparing_sem_videos}
\begin{figure}[!ht]
\centering 
\includegraphics[width=0.475\textwidth]{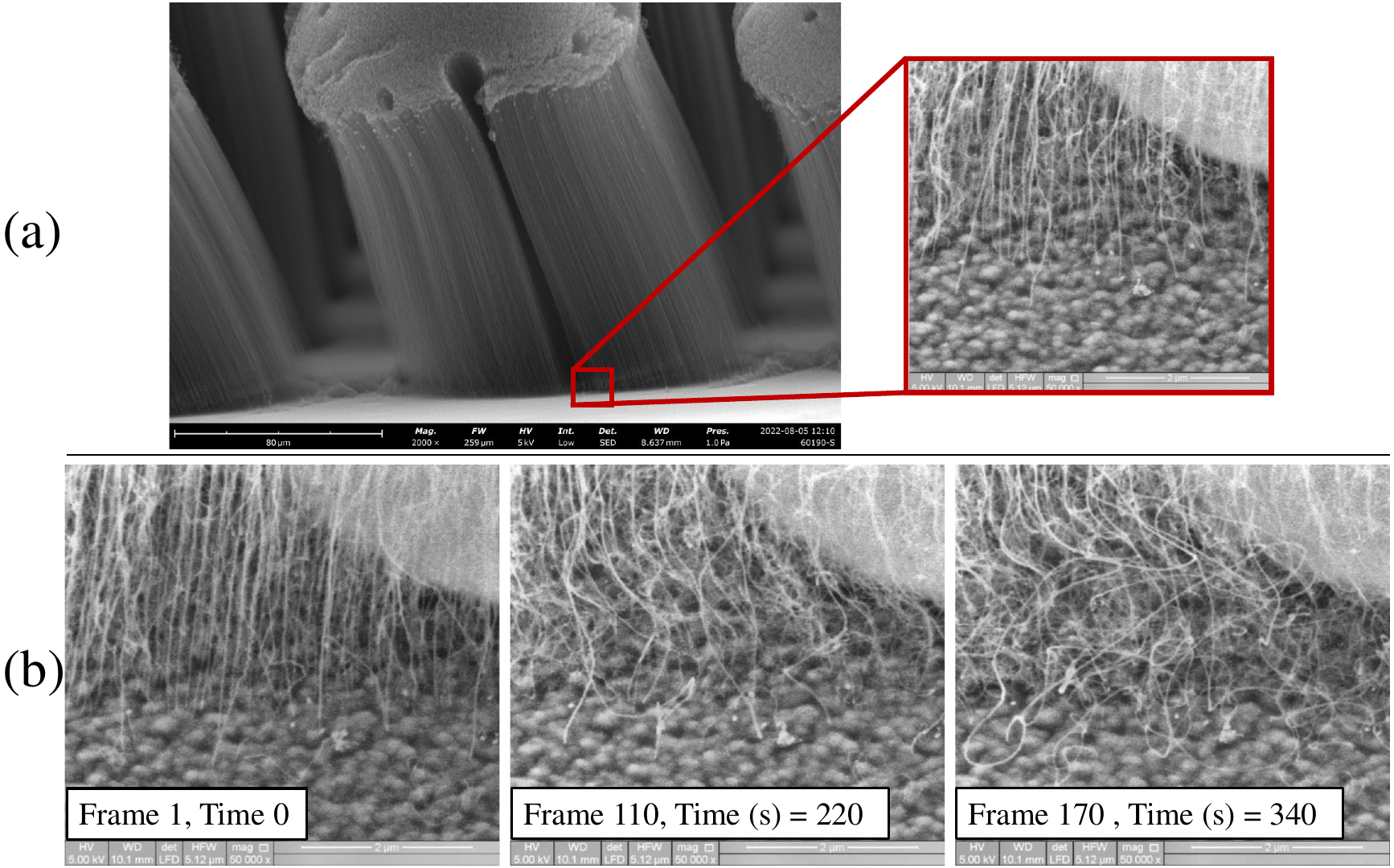}\\
\caption{(a) Target area for frame capturing in the SEM imaging process, and (b) sample SEM image sequence showing notable changes in CNT kinematics.}
\label{fig:cameraPositionGR}
\end{figure}

Carbon nanotubes were produced under a variety of growth conditions, encompassing alterations in temperature, substrate type, and growth methodologies, employing both fixed and floating catalyst chemical vapor deposition techniques. These methods enabled the creation of solid materials through gas-phase reactions, resulting in CNTs with unique mechanical properties influenced by the substrate and growth technique. The physical attributes of the carbon nanotubes were examined through desktop SEM imaging and nanoindentation tests. The SEM captures CNT images of the target area shown in \cref{fig:cameraPositionGR} (a) at one frame every 2 seconds, used to analyze CNT kinematics in this study. \Cref{fig:cameraPositionGR} (b) illustrates a sequence of SEM images showing significant changes in the kinematics of carbon nanotubes over time.
 
\subsection{Feature Tracking Algorithm}
\label{subsec:feature_tracking_algorithm}

The feature tracking \cite{gao2018evaluation, gao2019sensitivity} method aims to track and analyze keypoints movement across image sequences by identifying distinctive features in each frame, matching them, and recording their trajectories. The proposed algorithm integrates both traditional and contemporary techniques for feature detection and matching \cite{gao2020dct}. In this investigation, four distinct feature detection methods—SIFT, DISK, ALIKED, and SuperPoint—are coupled with the same matching method, LightGlue. The algorithm operates in five primary stages, as depicted in \cref{fig:trackingMethod} and elaborated in \cref{algo:featTracAlg}:

\begin{figure}[!ht]
  \centering 
  \includegraphics[width=.475\textwidth]{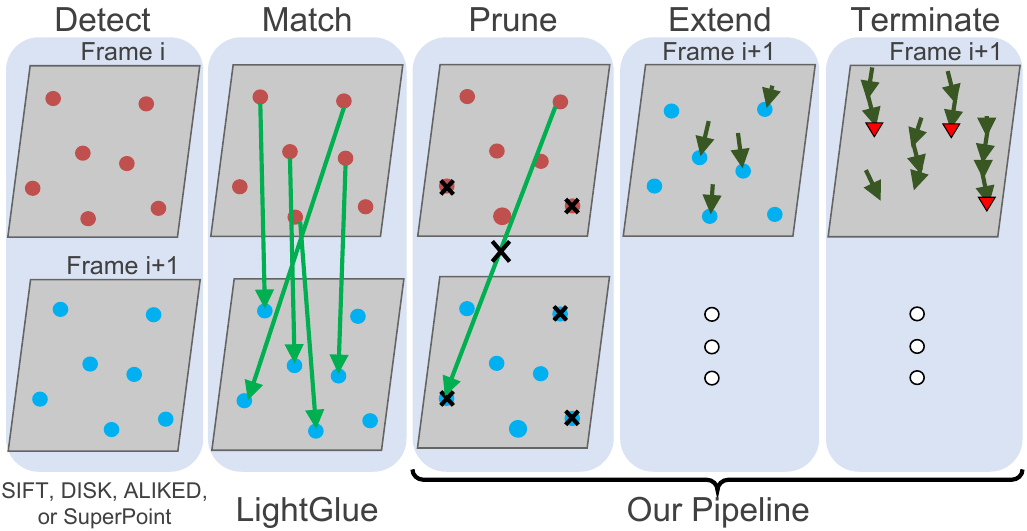}
  \caption{\textbf{Overview of VFTrack Framework:} \textbf{Detect:} Detection of features in subsequent frames using SIFT, DISK, ALIKED, or SuperPoint techniques. \textbf{Match:} Utilization of LightGlue for feature matching. \textbf{Prune:} Removal of keypoint mismatches and unpaired features. \textbf{Extend:} Formation of feature tracks and continue to grow as matched features are appended. \textbf{Terminate:} The algorithm considers tracks terminated (shown as red triangles) if no features match their heads and starts new tracks with the unmatched keypoints.}
\label{fig:trackingMethod}
\end{figure}

\begin{algorithm}[!ht]
  \caption{Feature Tracking Method}
  \label{algo:featTracAlg}
  \begin{algorithmic}[1]
    \REQUIRE frames $\mathit{frms}$, feature $\mathit{detector}$, feature $\mathit{matcher}$
    \ENSURE Tracks $\mathit{tracks}$, keypoints kinematic $\mathit{kptsKmts}$, track id $\mathit{trkID}$, and keypoints track id map $\mathit{kpTrkId}$
  
    \STATE Initialize $\mathit{tracks}, \mathit{prvKpts}, \mathit{kptsKmts}, \mathit{kpTrkId} \leftarrow \emptyset$
    \STATE $\mathit{kpts0} \leftarrow \mathit{detector}(\mathit{frm[0]})$, $\mathit{dist\_th} \leftarrow 40$, $\mathit{trkID} \leftarrow 0$ 

    \FOR{each $\mathit{kp} \in \mathit{kpts0}$}
      \STATE $\mathit{tracks[trkID]} \leftarrow [\mathit{kp}]$, $\mathit{kpTrkId[kp]} \leftarrow \mathit{trkID}$
      \STATE $\mathit{trkID} \gets + 1$, $\mathit{prvKpts} \gets \mathit{kpts0}$
    \ENDFOR

    \FOR{each $\mathit{(i, frm)} \in \mathit{frms[1:]}$}
      \STATE $\mathit{kpts} \leftarrow \mathit{\textbf{Detect}}(\mathit{frm})$

        \STATE $\mathit{mtchKpts} \leftarrow \mathit{\textbf{Match}}(\mathit{kpts}, \mathit{prvKpts})$
        
        \STATE $\mathit{mtchKpts} \leftarrow \text{\textbf{Prune}}(\mathit{mtchKpts}, \mathit{dist\_th})$
        \STATE $\mathit{tracks}, \mathit{kptsKmts}, \mathit{kpTrkId} \leftarrow $\par$ \text{\textbf{Extend}}(\mathit{mtchKpts}, \mathit{kpTrkId}, \mathit{tracks}, \mathit{kptsKmts})$
        \STATE $\mathit{kpTrkId}, \mathit{trkID} \gets $\par$ \text{\textbf{Terminate}}(\mathit{mtchKpts}, \mathit{kpTrkId}, \mathit{trkID})$
      \STATE $\mathit{prvKpts} \gets \mathit{kpts}$
    \ENDFOR
  \end{algorithmic}
\end{algorithm}

\begin{algorithm}[!ht]
  \caption{Prune Function}
  \label{algo:pruningAlg}
  \begin{algorithmic}[1]
    \REQUIRE $\mathit{mtchKpts}$, $\mathit{dist\_th}$
    \STATE \textbf{Function} $\text{\textbf{Prune}}(\mathit{mtchKpts}, \mathit{dist\_th})$
    \FOR{each $\mathit{(kp1, kp2)} \in \mathit{mtchKpts}$}
      \IF{$\text{Euclidean\_Distance}(kp1, kp2) > \mathit{dist\_th}$}
        \STATE \textbf{remove} $\mathit{(kp1, kp2)}$ \textbf{from} $\mathit{mtchKpts}$
      \ENDIF
    \ENDFOR
    \RETURN $\mathit{mtchKpts}$
  \end{algorithmic}
\end{algorithm}

\begin{algorithm}[!ht]
  \caption{Extend Function}
  \label{algo:ExtendAlg}
  \begin{algorithmic}[1]
    \REQUIRE $\mathit{mtchKpts}$, $\mathit{kpTrkId}$, $\mathit{tracks}$, $\mathit{kptsKmts}$
     
    \STATE \textbf{Function} $\text{\textbf{Extend}}(\mathit{mtchKpts}, \mathit{kpTrkId},$ \\ $\mathit{tracks}, \mathit{kptsKmts})$
    \FOR{each $\mathit{(kp1, kp2)} \in \mathit{mtchKpts}$}
      \STATE $\mathit{trkID} \leftarrow \mathit{kpTrkId[kp1]}$
      \STATE \textbf{append} $\mathit{(kp1, kp2)}$ \textbf{to} $\mathit{tracks[trkID]}$
      \STATE $\mathit{kptsKmts[trkID]} \leftarrow $\\$\text{calc\_kinematic}(\mathit{(kp1, kp2)}, \mathit{tracks[trkID]})$
      \STATE $\mathit{kpTrkId[kp2]} \leftarrow \mathit{trkID}$
    \ENDFOR
    \RETURN $\mathit{tracks}, \mathit{kptsKmts}, \mathit{kpTrkId}$
  \end{algorithmic}
\end{algorithm}

\begin{algorithm}[!ht]
  \caption{Terminate Function}
  \label{algo:terminateAlg}
  \begin{algorithmic}[1]
    \REQUIRE $\mathit{mtchKpts}$,  $\mathit{kpTrkId}$,  $\mathit{trkID}$

    \STATE \textbf{Function} $\text{\textbf{Terminate}}(\mathit{mtchKpts}, \mathit{kpTrkId}, \mathit{trkID})$
    \FOR{each $\mathit{(kp1, kp2)}$ \textbf{not in} $\mathit{mtchKpts}$}
      \STATE $\mathit{kpTrkId} \gets \mathit{kpTrkId} \setminus \{\mathit{kp1}\}$, $\mathit{trkID} \gets + 1$
      \STATE \textbf{append} $(\mathit{trkID}, \mathit{kp2})$ \textbf{to} $\mathit{kpTrkId}$
    \ENDFOR
    \RETURN $\mathit{kpTrkId}, \mathit{trkID}$
  \end{algorithmic}
\end{algorithm}

\noindent \textbf{Detect}: frame features were detected using the SIFT, DISK, ALIKED, or SuperPoint techniques (\cref{algo:featTracAlg}, line 11).\\
\textbf{Match}: Feature matching is performed between consecutive frames using LightGlue, chosen for its speed and precision (\cref{algo:featTracAlg}, line 12).\\
\textbf{Prune}: In this stage, the algorithm eliminates mismatches and unmatched features by enforcing a maximum displacement threshold between matched keypoints in consecutive frames. (\cref{algo:pruningAlg}).\\
\textbf{Extend}~\cite{kuchimanchi2025optimizing, 8639321}: Tracks form and grow by appending matched features to existing tracks if they match the features at the head of the tracks (\cref{algo:ExtendAlg}).\\
\textbf{Terminate}: Finally, the algorithm terminates tracks if no features match their heads and starts new tracks with the unmatched keypoints. (\cref{algo:terminateAlg}).

\subsection{Kinematic Decomposition of Motion Vectors}
\label{subsec:kinematic_decomposition}

The output of our visual feature tracking algorithm is a set of displacement vectors, one for each tracked CNT feature between consecutive frames. To extract meaningful physical insights, we decompose each raw displacement vector into components that represent distinct kinematic phenomena: axial growth, lateral drift, and oscillation.

\begin{figure}[!ht]
\centering 
\includegraphics[width=.30\textwidth]{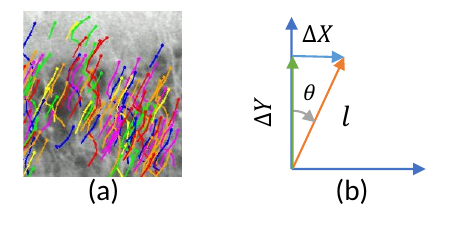}\\
\caption{Decomposition of a CNT feature's motion vector. (a) Raw displacement vectors tracked between frames. (b) Each velocity vector is resolved into a vertical component ($\Delta Y$), representing growth, and a horizontal component ($\Delta X$), representing drift and oscillation. The magnitude ($l$) and angle ($\theta$) are calculated from these components.}
\label{fig:Triangles}
\end{figure}

As illustrated in \Cref{fig:Triangles}, each displacement vector is resolved into a vertical component ($\Delta Y$) and a horizontal component ($\Delta X$) relative to the SEM image frame. We interpret the vertical displacement, $\Delta Y$, as the primary \textbf{growth} component along the pillar's main axis. The horizontal displacement, $\Delta X$, captures lateral movements, which we attribute to a combination of systemic \textbf{drift} of the entire pillar and localized \textbf{oscillations} of individual CNTs.

From these Cartesian components, we calculate the total displacement magnitude ($l$) and the instantaneous angle of motion ($\theta$) relative to the vertical growth axis using the following standard transformations:
\begin{align}
  l &= \sqrt{(\Delta X)^2 + (\Delta Y)^2} \\
  \theta &= \operatorname{atan2}(\Delta X, \Delta Y)
\end{align}
Here, $l$ represents the total distance a feature moved in a single time step, while $\theta$ quantifies its deviation from a purely vertical growth path. The temporal analysis of $\Delta Y$ across all tracks yields the average growth rate, while the distribution and evolution of $\theta$ for individual tracks are used to characterize CNT oscillations.

\subsection{Pillar Shape Estimation Methodology}
\label{subsec:pillar_shape_estimation_method}

\begin{figure}[!ht]
  \centering 
  \includegraphics[width=0.46\textwidth]{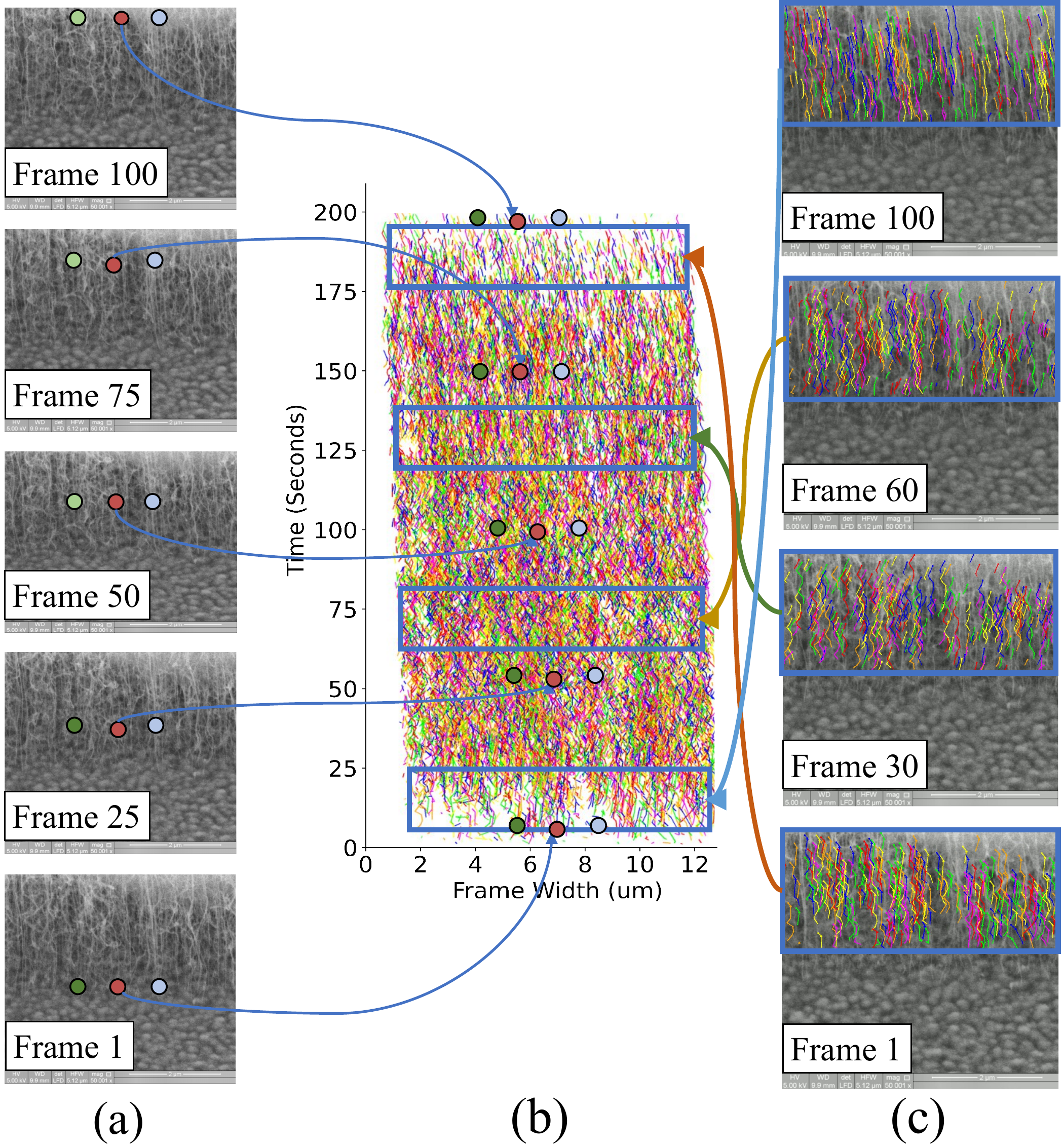}\\
  \caption{(a) CNT particles movement within the CNT micropillar over time. (b),(c): Visualizing the layer-by-layer reconstruction of a CNT micropillar. The method treats the collection of past tracks as a rigid body, which is shifted upward by the average growth and drift calculated for the current frame. The new tracks are then added to the base. This schematic shows the final positions of tracks from four sample frames (1, 30, 60, and 100) within a pillar constructed from a 100-frame sequence, mapping the temporal evolution into a final spatial structure.}~\label{fig:GrowthPillarSche}
\end{figure}

Estimating the three-dimensional morphology of carbon nanotube (CNT) micropillars is crucial for linking their structure to their mechanical and electrical properties. We propose a method to reconstruct the pillar shape by sequentially stacking the 2D CNT track data obtained from in-situ SEM video frames. The method operates on the key assumption that once nanotubes grow beyond the initial nucleation zone (\cref{fig:cameraPositionGR}), their collective structure behaves as a rigid body. In our iterative process, the algorithm first computes the average displacement vector (encompassing both growth and drift) from the tracked features in the current frame. This vector is then used to translate the entire previously reconstructed pillar shape. Finally, the new tracks from the current frame are superimposed at the pillar's base, representing the most recent growth. This layer-by-layer reconstruction is illustrated in \Cref{fig:GrowthPillarSche}, which shows a final pillar built from a 100-frame sequence. The figure highlights how tracks from earlier frames (e.g., Frame 1) form the top of the pillar, while tracks from the final frame (Frame 100) constitute the growing base.

\section{Experiment Results}
\label{sec:expermental_results}
This section validates the proposed feature tracking framework through kinematic analysis of CNT growth. We evaluate different feature detector-matcher pipelines to identify the optimal configuration for accuracy and real-time performance, then apply this pipeline to extract key physical parameters including spatially-resolved growth rates and orientation fluctuations. Finally, we demonstrate the framework's ability to reconstruct evolving 2D pillar shapes, providing comprehensive insight into growth dynamics.

\subsection{VFTrack Algorithm Performance Evaluation}
\label{subsec:Feature_Tracking_Analysis_Performance_Evaluation}
We employed four feature extraction methods—SIFT, DISK, ALIKED, and SuperPoint—coupled with LightGlue as the matcher. For this evaluation, a comprehensive ground-truth dataset was created by manually annotating 13,540 individual CNT trajectories across five distinct growth videos; this annotation was cross-referenced with results from the Meta CoTracker~\cite{karaev2024cotracker} to ensure high fidelity. The performance of each pipeline was rigorously evaluated using the standardized software from the Particle Tracking Challenge~\cite{chenouard2014objective}. While this framework provides several metrics (e.g., number of paired, generated, and missed tracks), we adopted the primary Jaccard similarity index, or $\alpha$ score, as the main indicator of tracking quality. The $\alpha$-score quantifies the similarity between the set of ground-truth tracks ($X$) and the estimated tracks ($Y$) using the relation:
\begin{align}
\alpha(X, Y) &= 1 - \frac{d(X, Y)}{d(X, \emptyset)}
\end{align}
where $d(X, Y)$ is the cumulative error distance between matched tracks, and the denominator $d(X, \emptyset)$ serves as a normalization factor based on the total length of the ground-truth tracks~\cite{chenouard2014objective}. The score ranges up to a maximum of 1, with a higher value indicating a closer approximation to the ground truth.

In addition, the following standard performance metrics were derived: precision, recall, and F1 score. These were computed using the formulas below:

\begin{align}
  \label{eq:precision}
  \text{Precision} &= \frac{\text{Paired Tracks (Alg., Ground Truth)}}{\text{Total Alg. Generated Tracks}}.
\end{align}

\begin{align}
  \label{eq:recall}
  \text{Recall} &= \frac{\text{Paired Tracks (Alg., Ground Truth)}}{\text{Total Ground Truth Tracks}}.
\end{align}

\begin{align}
  \label{eq:f1score}
  \text{F1 Score} &= \frac{2 \times \text{Precision} \times \text{Recall}}{\text{Precision} + \text{Recall}}.
\end{align}

The performance of each feature extraction method combined with either SuperGlue or LightGlue was evaluated using these metrics, alongside the runtime (seconds) per frame, as summarized in \cref{tab:method_metric_comparing}. The results clearly indicate that the LightGlue matcher provides superior accuracy and speed compared to SuperGlue across all detectors. Among the tested combinations, the \textbf{ALIKED + LightGlue} pipeline (highlighted in the table) emerged as the optimal choice, achieving the highest scores across all accuracy metrics, including an F1-score of 0.78 and an $\alpha$-score of 0.89. Although the SuperPoint + LightGlue pairing was faster, its accuracy was significantly lower. The ALIKED + LightGlue combination thus represents the best trade-off, delivering state-of-the-art tracking fidelity with a runtime (0.31s) well-suited for real-time applications. Consequently, this pipeline was adopted for all subsequent analyses in this study.

\begin{table}[!ht]
  \centering 
  {\small{
    \resizebox{0.47\textwidth}{!}{%
      \begin{tabular}{@{}lccccc@{}}
          \toprule
          Method            &Prec.$\uparrow$&Recall$\uparrow$ & F1$\uparrow$ & $\alpha$$\uparrow$ & Runtime$\downarrow$ \\ 
          \midrule
          SuperPoint + SuperGlue     & 0.57     & 0.59     & 0.55  & 0.54  & 0.54\\ 
          SIFT + SuperGlue           & 0.53     & 0.49     & 0.51  & 0.50  & 1.42\\ 
          DISK + SuperGlue           & 0.59     & 0.55     & 0.57  & 0.61 & 3.77\\ 
          ALIKED + SuperGlue          & 0.73     & 0.67     & 0.70  & 0.81 & 1.01\\ 
          \\
          \cmidrule{1-6}  
          
          SuperPoint + LightGlue     & 0.64     & 0.66     & 0.62  & 0.63  & \textbf{0.18}\\ 
          SIFT + LightGlue           & 0.60     & 0.56     & 0.58  & 0.58  & 0.55\\ 
          DISK + LightGlue           & 0.65     & 0.61     & 0.61  & 0.67 & 1.11\\ 
          \rowcolor{cyan!20}ALIKED + LightGlue & \textbf{0.81} & \textbf{0.75} & \textbf{0.78} & \textbf{0.89}  & 0.31\\ 
          \bottomrule
      \end{tabular}
    }
  }}
  \caption{Quantitative comparison of feature tracking pipelines, pairing four feature detectors with either the SuperGlue or LightGlue matcher. Performance is evaluated using precision, recall, F1-score, the $\alpha$-score (higher is better), and average runtime per frame in seconds (lower is better). The combination of ALIKED and LightGlue (highlighted) achieves the highest tracking accuracy, establishing it as the optimal method.}
  \label{tab:method_metric_comparing}
\end{table}

\begin{figure*}[!ht]
  \centering
  \begin{subfigure}[b]{0.246\textwidth}
      \includegraphics[width=\textwidth]{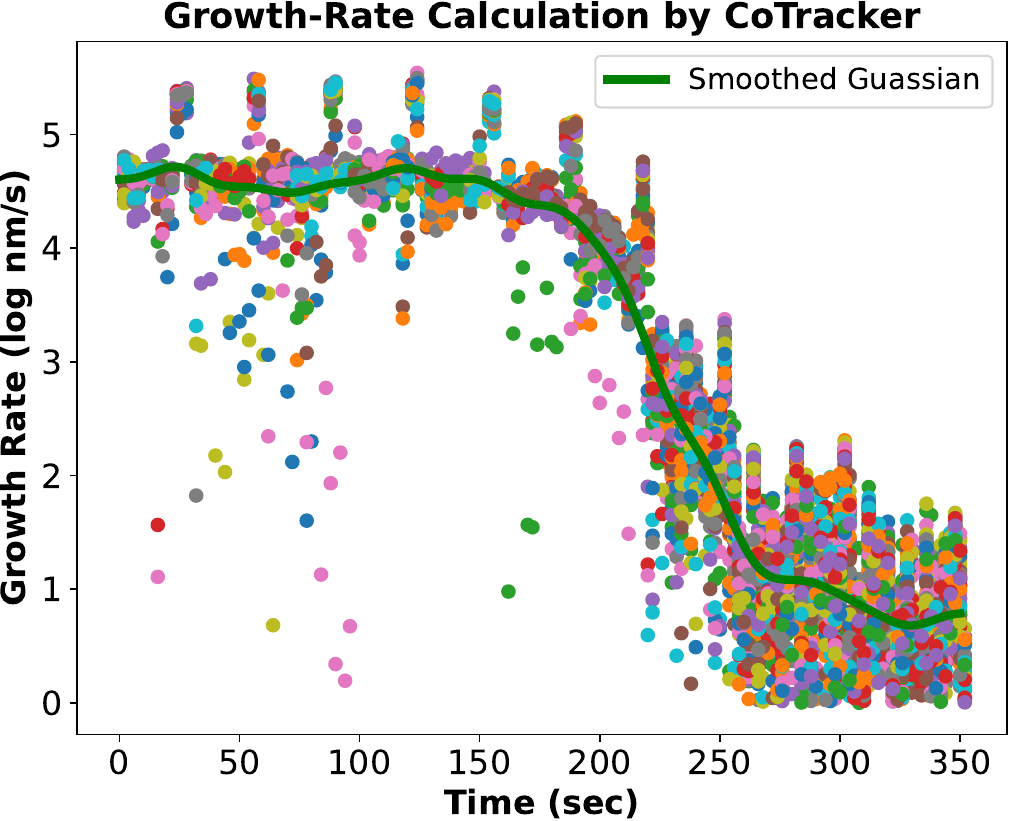}
      \caption{}
  \end{subfigure}
  \hfill
  \begin{subfigure}[b]{0.246\textwidth}
      \includegraphics[width=\textwidth]{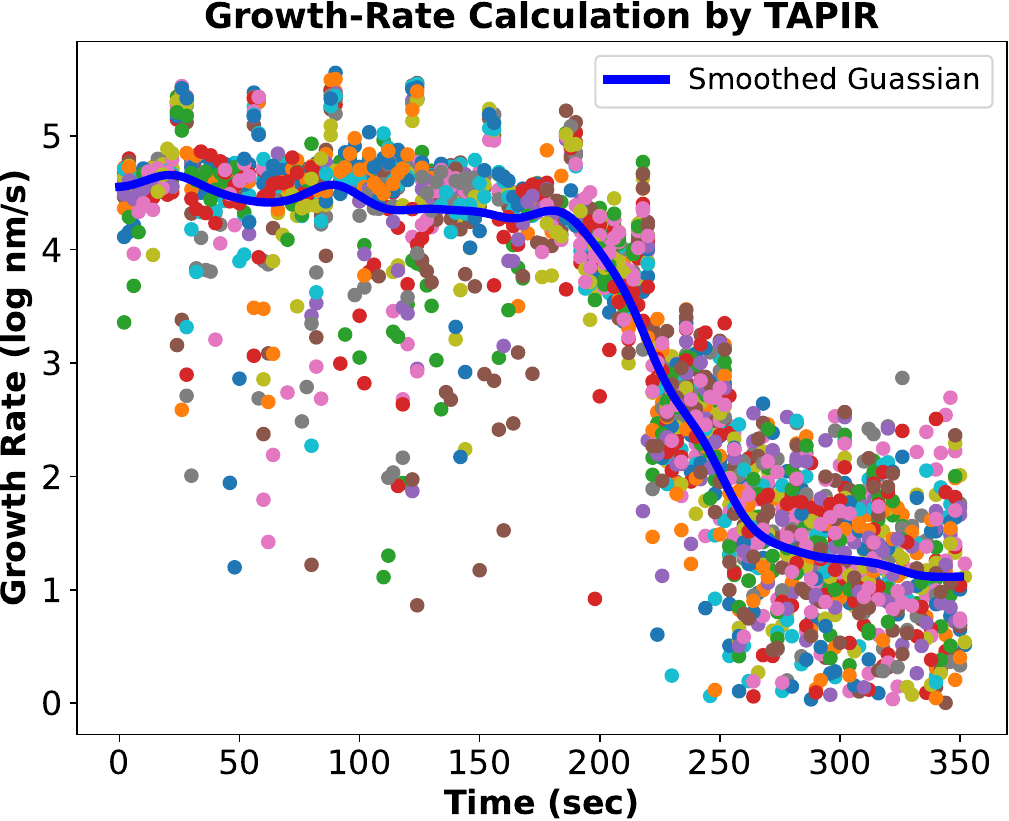}
      \caption{}
  \end{subfigure}
  \hfill
  \begin{subfigure}[b]{0.246\textwidth}
      \includegraphics[width=\textwidth]{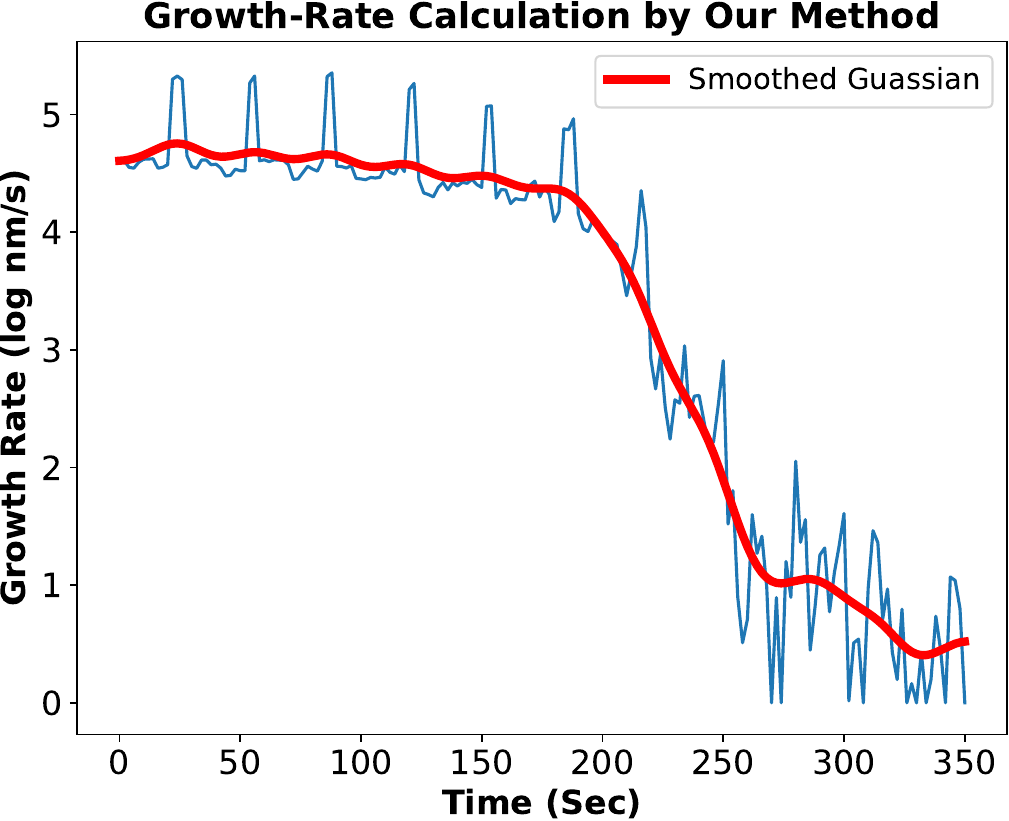}
      \caption{}
  \end{subfigure}
  \hfill
  \begin{subfigure}[b]{0.246\textwidth}
      \includegraphics[width=\textwidth]{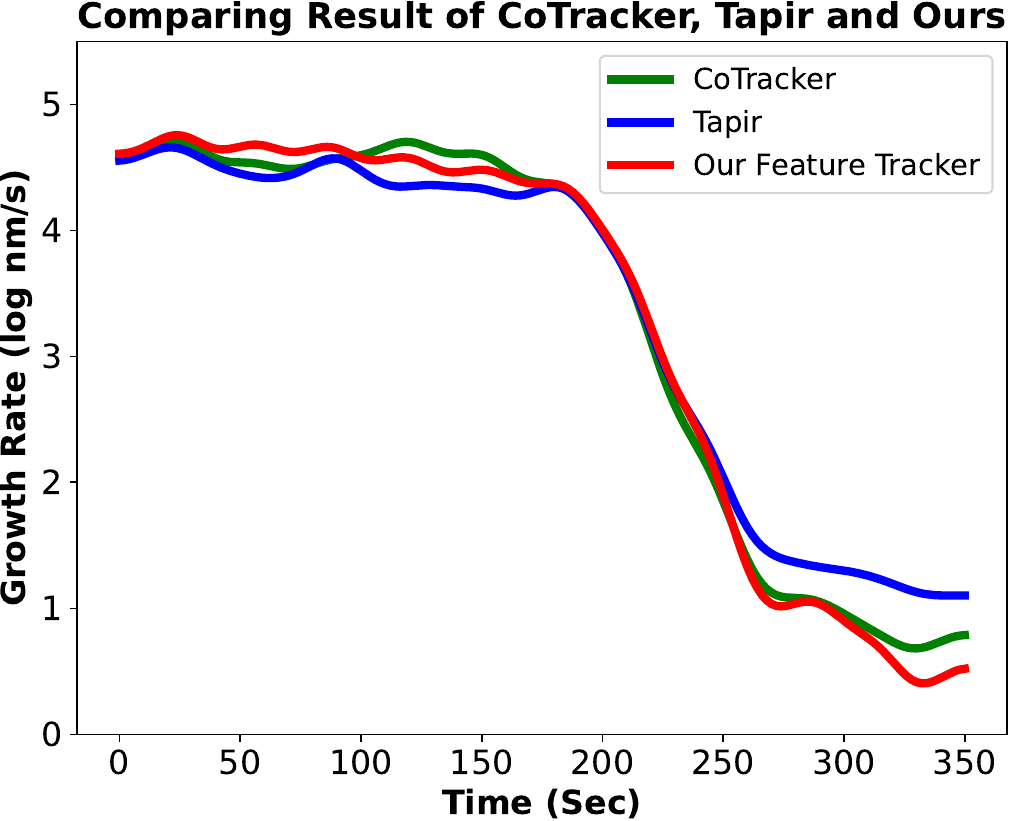}
      \caption{}
  \end{subfigure}
  \caption{(a) CNT growth rate calculation using CoTracker. (b) CNT growth rate calculation using TAPIR. (c) CNT growth rate calculation using VFTrack algorithm. (d) Comparison of growth rate calculations by different methods.}
  \label{fig:methods_GR_Calculation}
\end{figure*}

\begin{figure}[!ht]
  \centering
  \includegraphics[width=0.90\linewidth]{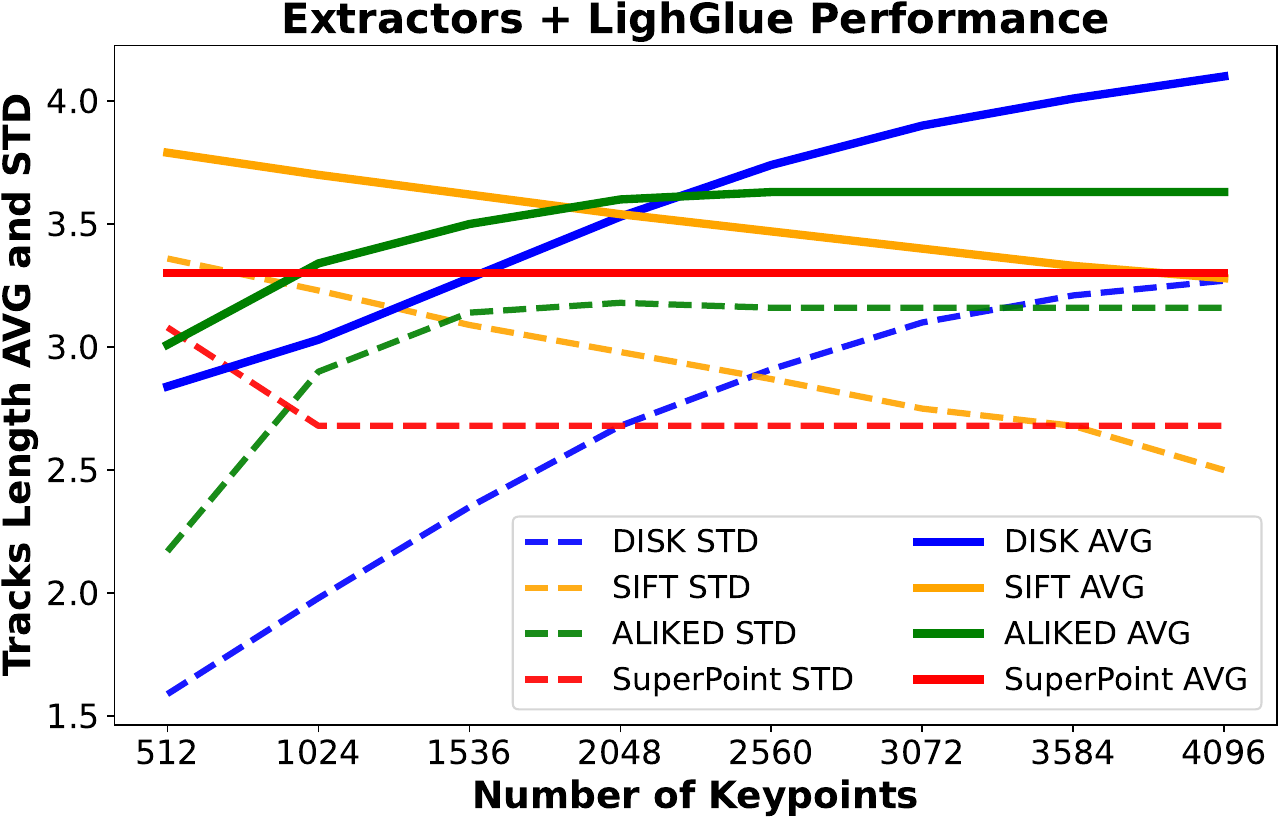}
  \caption{Comparison of average track length and standard deviation for feature tracking methods in SEM images. ALIKED + LightGlue shows the best overall performance, with SuperPoint + LightGlue excelling in low-density and DISK + LightGlue in high-density keypoint scenarios.}
  \label{fig:DetectorsPerformance}
\end{figure}

Additionally, the average length and standard deviation (STD) of the generated tracks were considered as metrics for comparing the variations of these methods. A higher track length with a lower STD indicates that the method can track CNT movement for an extended period. The results, shown in \cref{fig:DetectorsPerformance}, highlight that the SuperPoint and LightGlue combination performs well with a low number of keypoints per frame and maintains that performance even as the number of detected keypoints increases. Conversely, the DISK and LightGlue combination is the best option when dealing with a high number of keypoints per frame. The ALIKED+LightGlue combination offers the best balance and robustness in both low and high-density keypoint scenarios, making it versatile for a wide range of CNT dynamics in SEM imaging. Our analysis shows it consistently delivers reliable performance, balancing runtime efficiency with high tracking quality, making it the most effective approach across all feature densities.

\noindent \textbf{Real-time capability:} The proposed feature tracking algorithm demonstrates real-time compatibility for analyzing CNT kinematics in SEM microscopy, where SEM typically generates a frame every 2 seconds or less. As indicated in \cref{tab:method_metric_comparing}, our methods process each frame in under 1.1 seconds, even in the worst-case scenario (DISK + LightGlue). This means the algorithm is able to process a frame faster than the image acquisition rate of the SEM, ensuring seamless real-time operation. Furthermore, the SuperPoint + LightGlue method, with a processing time of 0.18 seconds per frame, stands out for ultra-fast real-time tracking applications.

\subsection{\textbf{CNT Growth Rate Calculation}}
\label{subsec:Feature_Tracking_Analysis}
The carbon nanotube growth rate, a vital parameter derived from our vector composition analysis in \cref{subsec:kinematic_decomposition}, was quantitatively evaluated using three distinct methods: two established point-tracking approaches (CoTracker \cite{karaev2024cotracker} and TAPIR \cite{doersch2023tapir}) and our novel visual feature-tracking algorithm. For CoTracker and TAPIR, growth rates were calculated from 206 manually selected keypoints across SEM frames (\cref{fig:methods_GR_Calculation}a-b), while our automated approach (\cref{fig:methods_GR_Calculation}c) autonomously identifies tracking points. \Cref{fig:methods_GR_Calculation}d demonstrates our method's superior growth rate estimation accuracy. Crucially, by eliminating manual keypoint initialization, our technique enables real-time CNT growth monitoring in SEM—a significant advancement over conventional point-tracking.

\begin{figure}[!ht]
  \centering 
  \includegraphics[width=.40\textwidth]{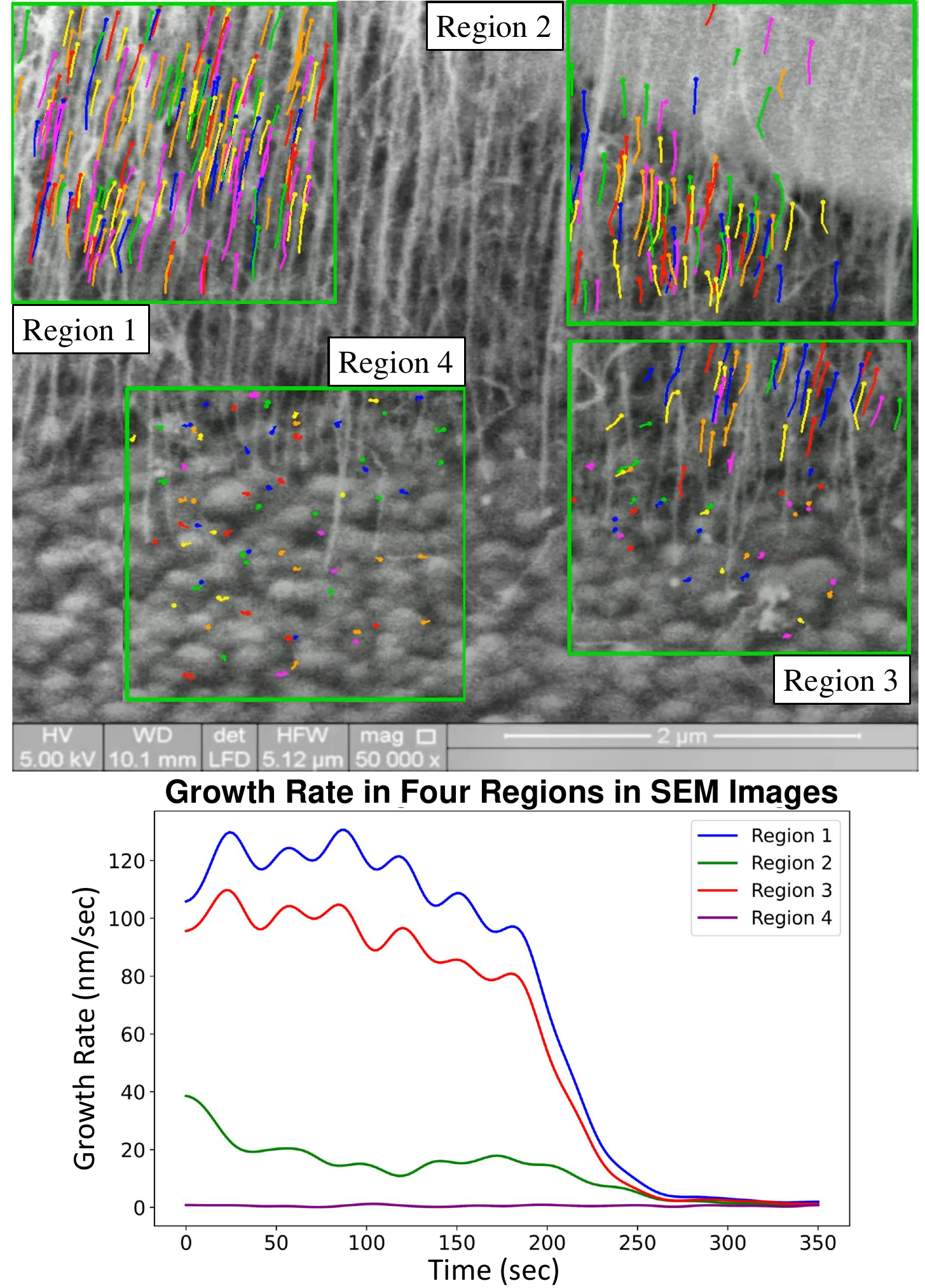}
  \caption{Spatially resolved growth rate analysis using our visual feature-tracking method. By defining four distinct regions, the algorithm reveals significant heterogeneity in the growth dynamics across the CNT pillar, a key advantage over global or sparse-point tracking methods.}
  \label{fig:gr_4_regions}
\end{figure}

\noindent \textbf{Regional Growth Rate Calculation:}
One of key advantage of the VFTrack method is its capacity to calculate growth rates within \textit{arbitrary regions} of SEM image sequences, as demonstrated in \cref{fig:gr_4_regions}. This capability captures complex CNT dynamics and evolving growth patterns across different sections of CNT micropillars over time.

\subsection{Temporal Analysis of CNT Tracks' Orientation}
\label{subsec:CNT_Orientation_Fluctuations}
\begin{figure}[!ht]
  \centering
  \includegraphics[width=0.90\linewidth]{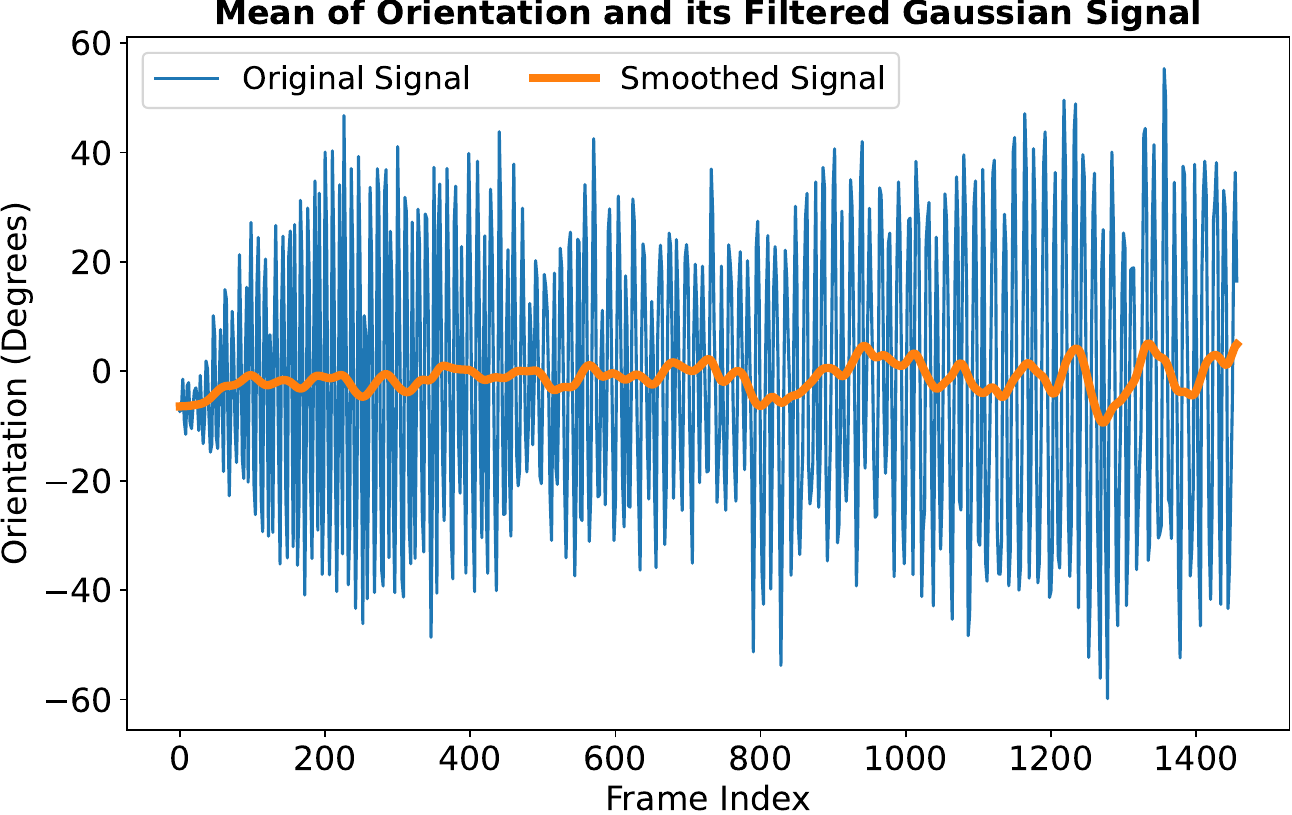}
  \caption{Mean orientation of carbon nanotubes over time, showing a general tendency to align vertically.}
  \label{fig:mean_orientation_gaussian}
\end{figure}

Building on the vector composition analysis in \cref{subsec:kinematic_decomposition}, the temporal evolution of Carbon Nanotube (CNT) orientation offers deeper insight into their dynamic behavior. By leveraging feature tracking, the orientation $\theta$ of CNT tracks are continuously monitored relative to a vertical reference. As shown in \cref{fig:mean_orientation_gaussian} the mean orientation of the tracked CNTs is observed to fluctuate around zero degrees, indicating that, on average, the CNTs tend to align vertically.

\begin{figure}[!ht]
  \centering
  \includegraphics[width=0.90\linewidth]{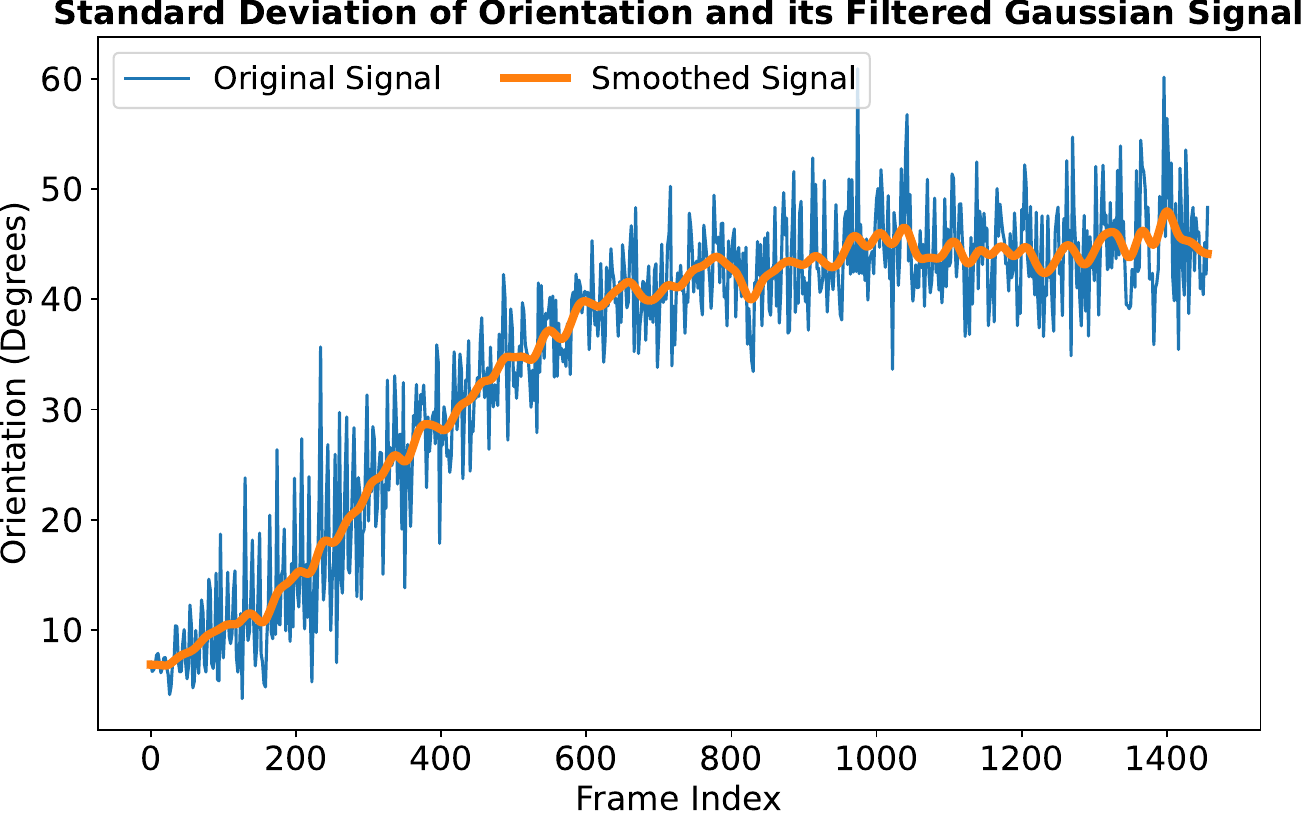}
  \caption{Increasing trend in the standard deviation of CNT orientations, rising from nearly zero to 50 degrees.}
  \label{fig:std_orientation_gaussian}
\end{figure}

However, the standard deviation of these orientations shows an increasing trend in fluctuation magnitude, as seen in \cref{fig:std_orientation_gaussian}. It rises from nearly zero, indicating minimal initial deviation, to about 50 degrees in the later frames.

The high fluctuation of CNTs and the decrease in their growth rate towards the end of the experiment can be attributed to several factors. Growing CNTs requires specific conditions, particularly the controlled flow of carbon-containing gas over a substrate at precise pressure and temperature. Initially, these conditions facilitate a high growth rate, forming a dense micropillar of CNTs. However, as the CNTs continue to grow and the pillar becomes taller, the increasing weight at the top creates a mechanical load and reduces the available space on the substrate for new CNTs to form. Additionally, variations in temperature, pressure, and carbonated gas flow at the substrate level become more pronounced, leading to the formation of fewer, thinner, and sparser CNTs at the base. This reduced density at the base, combined with the increased weight at the top, promotes the interconnection of CNTs, causing individual CNTs to deviate from their initial growth paths. These deviations result in significant fluctuations and oscillations in their structure, ultimately decreasing the overall growth rate and compromising the stability of the CNT structure.

\subsection{Pillar Shape Estimation Results}
\label{subsec:pillar_shape_estimation_results}
\begin{figure}[!ht]
  \centering 
  \includegraphics[width=0.445\textwidth]{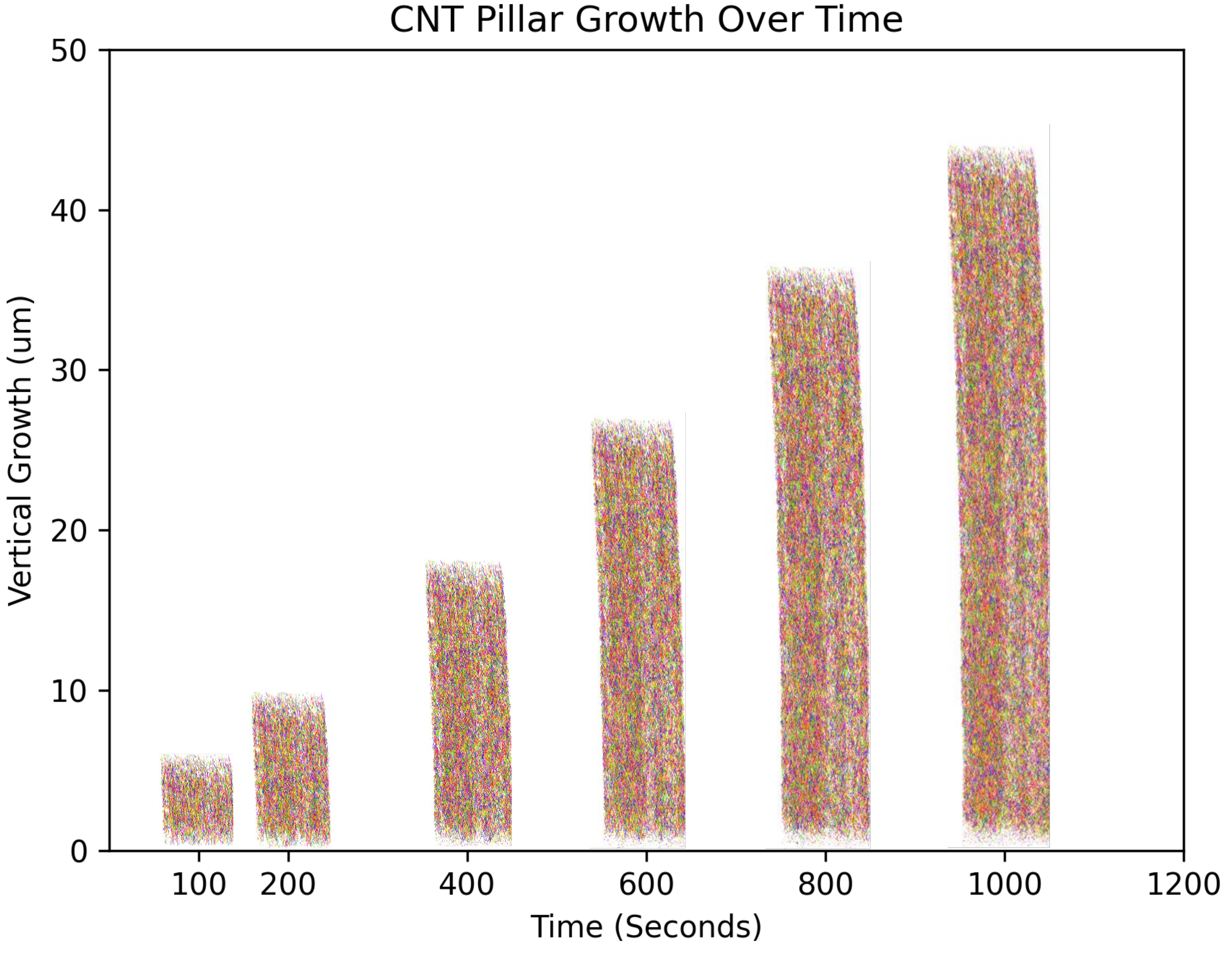}\\
  \caption{ Time-lapse visualization of a CNT micropillar’s expansion over 1000
seconds, showing steady increases in vertical growth.}
  \label{fig:PillarGrowthTime}
  \end{figure}

\begin{figure}[!ht]
  \centering 
  \includegraphics[width=0.44\textwidth]{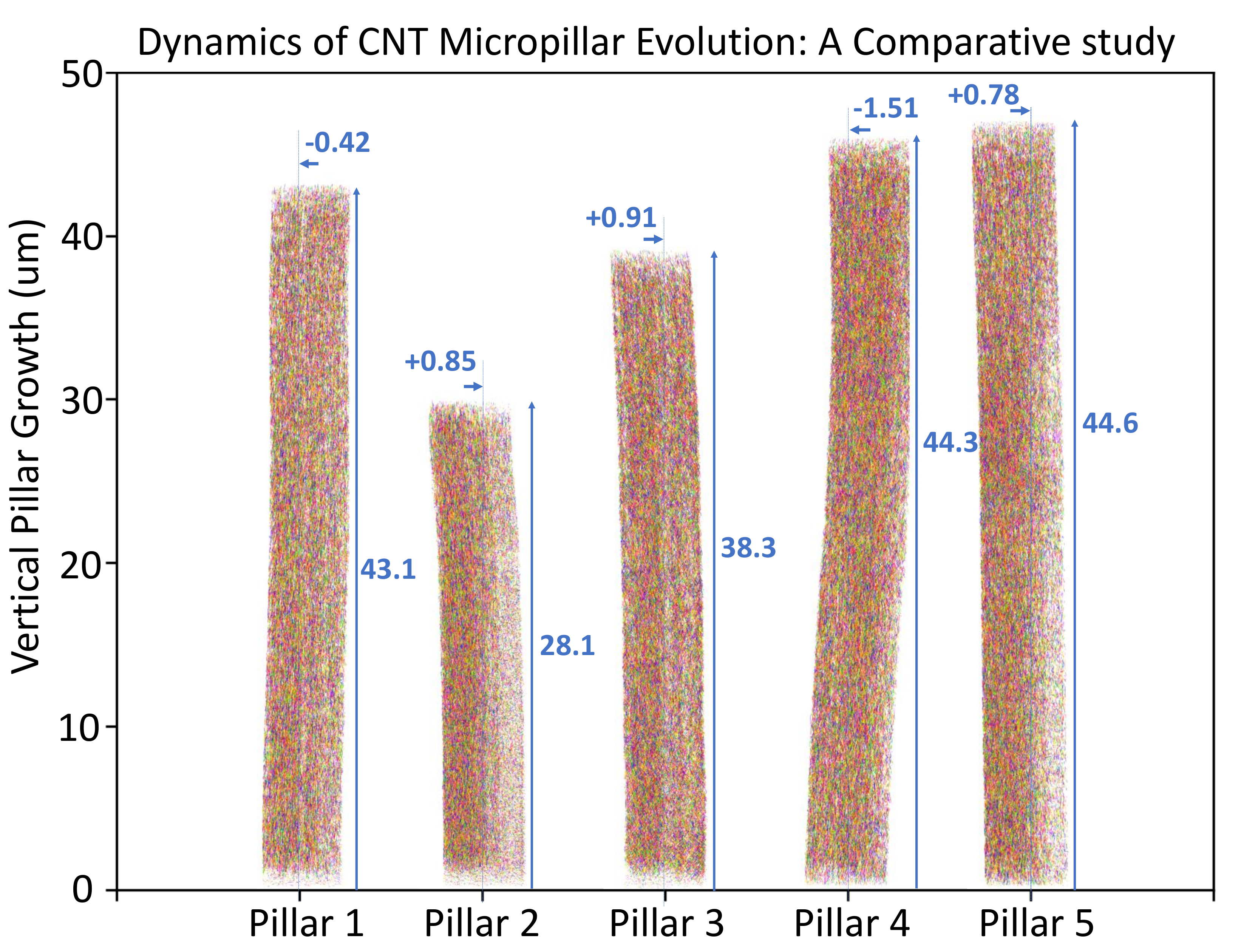}\\
  \caption{Comparative analysis of growth height and horizontal drift in CNT micropillars across five distinct image sequences.}
  \label{fig:PillarsComparing}
  \end{figure}

Methodology of pillar shape estimation was introduced in \cref{subsec:pillar_shape_estimation_method}. Figure \ref{fig:PillarGrowthTime}, demonstrates the incremental expansion of a single micropillar base within a CNT image sequence over 1000 seconds which reveals a steady increase in vertical growth. Same procedure was applied to five distinct image sequences, with the results illustrated in \cref{fig:PillarsComparing}. The figure illustrates both the growth height and the horizontal drift of the CNT micropillar, measured in micrometers. The results indicate that even minor changes in horizontal drift can significantly alter the final shape of the micropillar, potentially leading to substantial changes in the mechanical properties of the CNTs. The procedure was applied to five distinct image sequences, with the results illustrated in~\cref{fig:PillarsComparing}. The figure illustrates both the growth height and the horizontal drift of the CNT micropillar, measured in micrometers.

\begin{figure}[!ht]
  \centering
  \includegraphics[width=.45\textwidth]{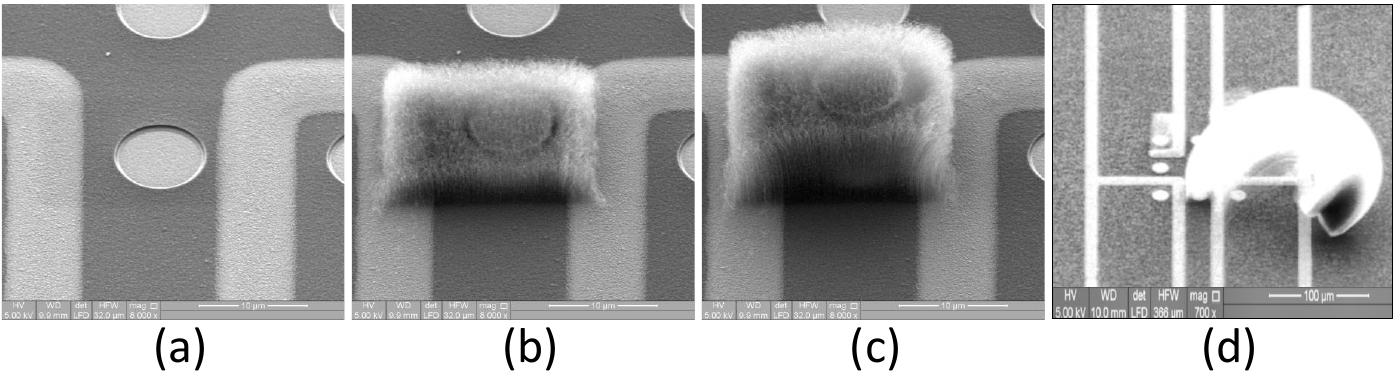}
  \caption{(a-c): Sequential images showing CNT micropillar's vertical growth. (d): Noticeable horizontal drift over time.}\label{fig:realCNT_growth}
\end{figure}

Both steady vertical growth and noticeable horizontal drift are observed in the real CNT micropillar as depicted in \cref{fig:realCNT_growth}. Images (a) through (c) illustrate the increasing height of the micropillar over time, while image (d) clearly shows the noticeable horizontal drift over time. The results demonstrate the effectiveness of the proposed method in estimating the shape of the CNT micropillar, providing valuable insights into the growth dynamics of CNTs in experimental settings.

\section{Conclusion and Future Works}
\label{sec:conclusion_future_works}\

In this work, we presented a novel, real-time feature-tracking framework designed to overcome the limitations of traditional ex situ and manual in situ methods for analyzing Carbon Nanotube (CNT) micropillar growth. Our comprehensive evaluation established the ALIKED detector paired with the LightGlue matcher as the optimal pipeline. By automatically detecting and tracking dense CNT particles, our method enables unprecedented kinematic analysis, including the decomposition of motion into growth, drift, and oscillation components, and the ability to compute regional growth rates. The successful reconstruction of evolving pillar morphologies, which captured both vertical growth and horizontal drift, further validates the framework's utility. 

The findings have significant implications for nanotechnology applications, particularly in fields requiring precise control of CNT behavior. Real-time monitoring of CNT growth can lead to more reliable flexible electronic components, optimized aerospace materials, and enhanced medical devices. Additionally, understanding CNT dynamics can improve the fabrication of CNT-based transistors, structural health monitoring systems, and environmental sensors.
    
However, integrating these insights into existing technologies presents challenges, including the need for significant computational resources and the high cost and time of real-time SEM experiments. Environmental factors like temperature and humidity also influence CNT behavior, requiring adaptive algorithms for accurate monitoring. Despite these challenges, our innovative methodology allows for a previously challenging level of detail precision in tracking CNT growth and behavior, providing dynamic insights into their mechanical properties. This knowledge can improve CNT-based products and lays a foundation for future research on environmental effects on CNT behavior, further advancing nanotechnology.
    
For future work, we plan to implement the algorithm on a local SEM machine for real-time data analysis, explore other feature extraction methods to enhance algorithm efficiency, and investigate the impact of environmental factors on CNT growth and movement. Combining advanced imaging with computational techniques, we anticipate further advancements in nanotechnology research and development of innovative CNT-based materials and devices.

\section*{Acknowledgments}\label{sec:acknowledgements}
All authors acknowledge the National Science Foundation grant CMMI-2026847 for funding. Any opinions, findings, and conclusions or recommendations expressed in this publication are those of the author(s) and do not necessarily reflect the views of the National Science Foundation.
{
    \small
    \bibliographystyle{ieeenat_fullname}
    \bibliography{main}
}

\end{document}